# A Tidy Data Model for Natural Language Processing using cleanNLP

*by Taylor Arnold*

**Abstract** Recent advances in natural language processing have produced libraries that extract low-level features from a collection of raw texts. These features, known as annotations, are usually stored internally in hierarchical, tree-based data structures. This paper proposes a data model to represent annotations as a collection of normalized relational data tables optimized for exploratory data analysis and predictive modeling. The R package **cleanNLP**, which calls one of two state of the art NLP libraries (CoreNLP or spaCy), is presented as an implementation of this data model. It takes raw text as an input and returns a list of normalized tables. Specific annotations provided include tokenization, part of speech tagging, named entity recognition, sentiment analysis, dependency parsing, coreference resolution, and word embeddings. The package currently supports input text in English, German, French, and Spanish.

## Introduction

There has been an ongoing trend towards converting raw data into a collection of normalized tables prior to conducting further analyses. This paradigm, recently popularized by Hadley Wickham under the term "tidy data" (Wickham, 2014), draws on concepts from database and visualization theory to provide a welcomed theoretical basis for data analysis. There are also many pragmatic benefits to putting data into a set of normalized tables prior to beginning an exploratory analysis or building inferential models. When working with normalized data most modeling, data manipulation, and visualization tasks can be described using a small collection of functions. This makes code more readable, less-error prone, and allows for better code reuse. As many of these simple functions reduce to basic database operations, this style of coding can simplify the task of integrating statistical models into a production codebase. Also, normalized tables can be stored unambiguously as delimited plain text flat files, allowing for interoperability between programming languages and users.

As both a cause and result of the popularity of this approach, a number of software packages have been developed to help construct and manipulate collections of normalized data tables. In R, well-known examples include **dplyr** (Wickham and Francois, 2016), **ggplot2** (Wickham, 2009), **magrittr** (Bache and Wickham, 2014), **broom** (Robinson, 2017), **janitor** (Firke, 2016), and **tidyr** (Wickham, 2017). On the Python side, much of this functionality is included within the pandas (McKinney et al., 2010) and sklearn (Pedregosa et al., 2011) modules.

While cleaning messy data is often a time-consuming task, deciding on a specific normalized schema for representing a set of inputs is in most cases relatively straightforward. Outside of potentially removing outliers, missing data, and bad inputs, the process of tidying data is generally a lossless procedure. At a high-level, data tidying is often simply a reorganization of the raw inputs. However, if we are working with unstructured data such as collections of text, images, or sound, converting into a normalized tabular format is significantly more involved. The process of tidying in these cases becomes synonymous with featurization, whereby structured outputs are algorithmically extracted from a raw input. For example, from an audio music file we might extract features such as the overall length, beats per minute, and quantiles of the music's loudness.

The featurization of raw text, known in natural language processing as *text annotation*, includes tasks such as tokenization (splitting text into words), part of speech tagging, and named entity recognition. Recent advancements in neural networks and heavy investment from both industry and academia have produced fast and highly accurate annotation libraries such as Stanford's CoreNLP (Manning et al., 2014), spaCy (Honnibal and Johnson, 2015), Apache's OpenNLP (Baldridge, 2005), and Google's SyntaxNet (Petrov, 2016). All of these, however, internally represent annotations using collections of complex, hierarchical, object-oriented classes. While these structures are ideal for annotation, they are not optimal for exploratory and predictive modeling.

In this paper, we present a method for uniting the cutting edge advancements in natural language processing with the popular normalized data paradigm. Specifically, we give a data schema representing the output of an NLP annotation pipeline as a collection of normalized tables. Alongside this specification, we present the R package **cleanNLP** that implements this specification over three distinct back ends. The package contains:

- custom Java code, called by **rJava** (Urbanek, 2016), that annotates raw text using the CoreNLP library;

- a custom Python script, called by **reticulate** (Allaire et al., 2017), that annotates raw text using the spaCy library;

- a simple, system dependency free, annotation engine using the package **tokenizers** (Mullen, 2016).

The package **cleanNLP** also includes tools for converting from the normalized data model into (sparse) data matrices appropriate for exploratory and predictive modeling. Together, these contributions simplify the process of doing exploratory data analysis over a corpus of text.

There are several existing R packages that have some similar or complementary features to those in **cleanNLP**. The R package **tidytext** (Silge and Robinson, 2016) also offers the ability to convert raw rext into a data frame. It is quite similar to the functionality of **cleanNLP** when using the **tokenizers** back end, with the addition of basic sentiment analysis and part of speech tagging for English through the use of word lists. With all annotations occurring at the token level, results are given as a single table rather than a normalized schema between many tables as in **cleanNLP**, which simplifies its application for new users. As such, **tidytext** works well for applications that do not need more advanced annotators such as named entities, dependencies, and coreferences. Given the overlap in general approaches, it should be relatively straightforward for users to transition from **tidytext** to **cleanNLP** when they find the need for these annotation tasks. There are two existing R packages that also call functions in the CoreNLP library. The package **StanfordCoreNLP** (Hornik, 2016c), available only through the datacube website at Vienna University, integrates into the **NLP** framework. A similar, standalone approach is offered by **coreNLP** (Arnold and Tilton, 2016). Both of these packages run the annotation pipeline over a corpus of text, call the java class edu.stanford.nlp.pipeline.XMLOutputter, and then parse the output using the XML package. This approach is not ideal as parsing the output XML file is computationally time-consuming. It is also error prone because there is no published format specifying the output of the XML.[1] There is also the package **spacyr** (Benoit and Matsuo, 2017), which was published after **cleanNLP**, that offers another way of calling the spaCy library from R. Internally, **spacyr** works similarly to the spaCy back end in **cleanNLP** by calling the Python library and extracting information into R data types. However, **spacyr** returns results as a single denormalized data frame and (perhaps in part as a result of having no easy way of storing them in the one-table output) does not support the word embeddings feature of the spaCy library.

The package has been designed to integrate into workflows that utilize the many other packages for text processing available in R, such as those found in the CRAN Taskview *NaturalLanguageProcessing*. For example, users may use the framework provided by **tm** (Feinerer et al., 2008) to manage external corpora or the classes within NLP (Hornik, 2016a) to run alternative parsers that can be converted into a tidy framework by way of the from_CoNLL function. The Apache OpenNLP annotation pipeline, available via **openNLP** (Hornik, 2016b), for instance, provides several languages not yet supported by spaCy or the CoreNLP pipeline. Packages that focus on the analysis and modeling of text data can usually be used directly with the output from cleanNLP; these include **lda** (Chang, 2015), **lsa** (Wild, 2015), and **topicmodels** (Grün and Hornik, 2011). Similarly, general-purpose database back-ends such as **sqliter** (Freitas, 2014) can be used to store the tidy data tables; predictive modeling functions may be used to do predictive analytics over generated term-frequency matrices.

In the following section we illustrate the usage of the R package across all three back ends. Next, we give a detailed description and justification of our data model. Along the way, we give a high-level introduction to the ideas behind the underlying NLP annotators. We finish by illustrating a longer example of using the package to study a corpus of historical speeches made by Presidents of the United States.

## Basic usage of cleanNLP

Before describing the data model for text annotations, it is useful to understand the basic workflow provided by the R package **cleanNLP**. We start by writing the opening lines of Douglas Adams' *Life, the Universe and Everything* to a temporary file.

```
> txt <- c("The regular early morning yell of horror was the sound of Authur",
+          "Dent waking up and suddenly remembering where he was. It wasn't",
+          "just that the cave was cold, it wasn't just that it was damp and",
+          "smelly. It was the fact that the cave was in the middle of",
```

---

[1] This author, who is also the maintainer of **coreNLP**, has witnessed this first-hand by way of the persistent bug reports centering around the formatting of the XML output in strange edge cases or over new versions of the CoreNLP library. The **coreNLP** will still be maintained for users looking explicitly to access methods from the Stanford Library, whereas **cleanNLP** is being developed to provide a simpler interface that is consistent across various back ends.

```
+              "Islington and there wasn't a bus due for two million years.")
> writeLines(txt, tf <- tempfile())
```

The package **cleanNLP** can be installed directly from CRAN, with binaries available for all major operating systems. In order to annotate raw text, an NLP back end must first be initalized. Once this is done, annotation is done by calling the function `annotate` with a vector of path(s) to the input documents. We start with an example using the tokenizers back end.

```
> library(cleanNLP)
> init_tokenizers()
> anno <- run_annotators(tf)
```

The result of the annotation is a named list of six data frames and one matrix. We can see the elements of the object by printing out their names.

```
> names(anno)
[1] "coreference" "dependency"  "document"    "entity"      "sentence"
[6] "token"       "vector"
```

The individual tables can be referenced with the generic R accessor functions (such as `` `[[` ``), however the preferred method is to call the relevant **cleanNLP** functions of the form `get_TABLENAME()`. For example, the tokens table for this example can be accessed with the `get_token` function.

```
> get_token(anno)
# A tibble: 61 x 8
      id   sid   tid    word lemma  upos   pos   cid
   <int> <int> <int>   <chr> <chr> <chr> <chr> <int>
1      1     0     1     The  <NA>  <NA>  <NA>    NA
2      1     0     2 regular  <NA>  <NA>  <NA>    NA
3      1     0     3   early  <NA>  <NA>  <NA>    NA
4      1     0     4 morning  <NA>  <NA>  <NA>    NA
5      1     0     5    yell  <NA>  <NA>  <NA>    NA
6      1     0     6      of  <NA>  <NA>  <NA>    NA
7      1     0     7  horror  <NA>  <NA>  <NA>    NA
8      1     0     8     was  <NA>  <NA>  <NA>    NA
9      1     0     9     the  <NA>  <NA>  <NA>    NA
10     1     0    10   sound  <NA>  <NA>  <NA>    NA
# ... with 51 more rows
```

The get functions are preferable because they provide useful options for modifying the output before returning it. Notice that the annotation process here has split out each word in the input into its own row. There are also several columns of ids and columns filled with missing values. The specific schema of the tables will be the focus of discussion in the following section.

The tokenizers back end requires no external dependencies, however it does not support any of the advanced annotation tasks that illustrate the utility of the **cleanNLP** package. This explains why most of the columns in the example are missing. It is included primarily for testing and demonstration purposes in cases where the other back ends cannot be installed. The spaCy back end uses the Python library by the same name for the purpose of extracting text annotations. Users must install Python and the library externally (detailed instructions are provided in the package documentation). Once installed, the only modification required by the R code is to adjust which `init_` function is being called.

```
> init_spaCy()
> anno <- run_annotators(tf)
> get_token(anno)
# A tibble: 68 x 8
      id   sid   tid    word   lemma  upos   pos   cid
   <int> <int> <int>   <chr>   <chr> <chr> <chr> <int>
1      1     1     1     The     the   DET    DT     0
2      1     1     2 regular regular   ADJ    JJ     4
3      1     1     3   early   early   ADJ    JJ    12
4      1     1     4 morning morning  NOUN    NN    18
5      1     1     5    yell    yell  NOUN    NN    26
6      1     1     6      of      of   ADP    IN    31
7      1     1     7  horror  horror  NOUN    NN    34
8      1     1     8     was      be  VERB   VBD    41
9      1     1     9     the     the   DET    DT    45
```

| table name | record | primary key | foreign keys |
|------------|--------|-------------|--------------|
| document | document | id | · |
| token | word / punctuation | id, sid, tid | cid |
| dependencies | token pairs | id, sid, tid, tid_target | · |
| entity | set of tokens | id, sid, tid, tid_end | · |
| coreference | mentions | id, rid, mid | sid, tid, tid_end, tid_head |
| sentence | sentence | id, sid | · |
| vector | word embedding | id, sid, tid | · |

**Table 1:** Tables in the data model and their (composite) primary and foreign keys. All keys are given by non-negative integers. Namely, `id` indexes the documents, `sid` the sentences within a document, and `tid` the tokens within a sentence. The `cid` gives character offsets into the raw input text. Keys `rid` and `mid` are specifically constructed by the coreference annotator.

```
10     1     1    10    sound    sound  NOUN    NN    49
# ... with 58 more rows
```

The output is in the exact some format but now all of the token columns are filled in with useful information such as the lemmatized form of each word and part of speech codes. Similar details are also filled into the other fields.

The third and final back end currently available uses the Java library coreNLP. Users must install Java version 1.8 or higher and link it to R using the rJava. The coreNLP models, which are over 1 GB, can then be either manually downloaded or grabbed using the helper function `download_coreNLP()`. Once installed, the back end works just as with the other back ends.

```
> init_coreNLP()
> anno <- run_annotators(tf)
> get_token(anno)
# A tibble: 68 x 8
      id   sid   tid    word   lemma  upos   pos   cid
   <int> <int> <int>   <chr>   <chr> <chr> <chr> <int>
1      1     1     1     The     the   DET    DT     0
2      1     1     2 regular regular   ADJ    JJ     4
3      1     1     3   early   early   ADJ    JJ    12
4      1     1     4 morning morning  NOUN    NN    18
5      1     1     5    yell    yell  VERB    VB    26
6      1     1     6      of      of   ADP    IN    31
7      1     1     7  horror  horror  NOUN    NN    34
8      1     1     8     was      be  VERB   VBD    41
9      1     1     9     the     the   DET    DT    45
10     1     1    10   sound   sound  NOUN    NN    49
# ... with 58 more rows
```

The token output here is similar, but not exactly the same, as that produced by the spaCy annotation engine. The only distinction in the first ten rows is whether the word *yell* is categorized as a noun (spaCy) or a verb (coreNLP). While yell can be either part of speech, in context the spaCy interpretation is correct.

As seen in the code-snippets here, the philosophy behind the design of the **cleanNLP** package is to make it as easy as possible to get raw text turned into data frames. All of the functions introduced here have optional parameters that change the way the back ends are run or how the annotations are returned. This includes which annotators to run and selecting the desired language model to use. Complete documentation is available within the R help pages.

## A data model for the NLP pipeline

An annotation object is simply a named list with each item containing a data frame. These frames should be thought of as tables living inside of a single database, with keys linking each table to one another. All tables are in the second normal form of Codd (1990). For the most part they also satisfy the third normal form, or, equivalently, the formal tidy data model of Wickham (2014). The limited departures from this more stringent requirement are justified below wherever they exist. In every case the cause is a transitive dependency that would require a complex range join to reconstruct.

Several standards have previously been proposed for representing textual annotations. These include the linguistic Annotation Framework (Ide and Romary, 2001), NLP Interchange Format (Hellmann et al., 2012), and CoNLL-X (Buchholz and Marsi, 2006). The function from_CoNLL is included as a helper function in **cleanNLP** to convert from CoNLL formats into the **cleanNLP** data model. All of these, however, are concerned with representing annotations for interoperability between systems. Our goal is instead to create a data model well-suited to direct analysis, and therefore requires a new approach.

| | get_document() |
|---|---|
| id | integer. Id of the source document. |
| time | date time. The time at which the parser was run on the text. |
| version | character. Version of the NLP library used to parse the text. |
| language | character. Language of the text, in ISO 639-1 format. |
| uri | character. Description of the raw text location. |

**Table 2:** Schema for the document table. The id field serves as a primary key, and other meta data fields may be appended that give domain-specific information about each document.

In this section each table is presented and justifications for its existence and form are given. Individual tables may be pulled out with access functions of the form get_*. Example tables are pulled from the dataset obama, which is included with the **cleanNLP** package. This gives the annotation object obtained from the text of the annual speeches Barack Obama made to Congress. These annual addresses, known as *The State of the Union*, are mandated by the US Constitution and have been given by every president since George Washington.

### Documents

The documents table contains one row per document in the annotation object. What exactly constitutes a document is up to the user. It might include something as granular as a paragraph or as coarse as an entire novel. For many applications, particularly stylometry, it may be useful to simultaneously work with several hierarchical levels: sections, chapters, and an entire body of work. The solution in these cases is to define a document as the smallest unit of measurement, denoting the higher-level structures as metadata. For example, when working with a corpus of texts where each book is broken into chapters, we would make each document an individual chapter. A metadata field would be assigned to each chapter indicating which book it is a part of.

The primary key for the document table is a document id, stored as an integer index. By design, there should be no extrinsic meaning placed on this key. Other tables use it to map to one another and to the document table, but any metadata *about* the document is contained only in the document table rather than being forced into the document key. In other words, the temptation to use keys such as "Obama2016" is avoided because, while these look nice, trying to make use of them to extract document-level metadata is error prone and ultimately more verbose than making use of a join with the document table.

The minimal fields required by the document table are given in Table 2. These are all filled in automatically by the annotation function. Any number of additional corpora-specific metadata, such as the aforementioned section and chapter designations, may be attached as well by giving it as an option to the meta parameter of run_annotators. The document table for the example corpus is:

```
> get_document(obama)
# A tibble: 8 x 5
     id                time version language      uri
  <int>              <dttm>   <chr>    <chr>    <chr>
1     1 2017-05-21 09:27:55   1.8.2       en 2009.txt
2     2 2017-05-21 09:28:00   1.8.2       en 2010.txt
3     3 2017-05-21 09:28:05   1.8.2       en 2011.txt
4     4 2017-05-21 09:28:10   1.8.2       en 2012.txt
5     5 2017-05-21 09:28:14   1.8.2       en 2013.txt
6     6 2017-05-21 09:28:18   1.8.2       en 2014.txt
7     7 2017-05-21 09:28:22   1.8.2       en 2015.txt
8     8 2017-05-21 09:28:26   1.8.2       en 2016.txt
```

It may seem that common fields such as year and author should be added to the formal specification but the perceived advantage is minimal. It would still be necessary for users to manually add the content of these fields at some point as any other metadata is not unambiguously extractable from the raw text.

|       | get_token()                                                                           |
|-------|---------------------------------------------------------------------------------------|
| id    | integer. Id of the source document.                                                   |
| sid   | integer. Sentence id, starting from 0.                                                |
| tid   | integer. Token id, with the root of the sentence starting at 0.                       |
| word  | character. Raw word in the input text.                                                |
| lemma | character. Lemmatized form the token.                                                 |
| upos  | character. Universal part of speech code.                                             |
| pos   | character. Language-specific part of speech code; uses the Penn Treebank codes.       |
| cid   | integer. Character offset at the start of the word in the original document.          |

**Table 3:** Schema for the token table. The fields id, sid, and tid serve as a composite key for each token. A row also exist for the root of each sentence.

### Tokens

The token table contains one row for each unique token, usually a word or punctuation mark, in any document in the corpus. Any annotator that produces an output for each token has its results displayed here. These include the lemmatizer and the part of the speech tagger (Toutanova and Manning, 2000). Table 3 shows the required columns contained in the token table. Given the annotators selected during the pipeline initialization, some of these columns may contain only missing data. A composite key exists by taking together the document id, sentence id, and token id. There is also a foreign key, `cid`, giving the character offset back into the original source document. An example of the table looks like this:

```
> get_token(obama, include_root = TRUE)
# A tibble: 65,758 x 8
      id   sid   tid      word     lemma   upos    pos    cid
   <int> <int> <int>     <chr>     <chr>  <chr>  <chr>  <int>
1      1     1     0      ROOT      ROOT   <NA>   <NA>     NA
2      1     1     1     Madam     madam  PROPN    NNP      0
3      1     1     2   Speaker   speaker  PROPN    NNP      6
4      1     1     3         ,         ,  PUNCT      ,     13
5      1     1     4       Mr.       mr.  PROPN    NNP     15
6      1     1     5      Vice      vice  PROPN    NNP     19
7      1     1     6 President president  PROPN    NNP     24
8      1     1     7         ,         ,  PUNCT      ,     33
9      1     1     8   Members   members  PROPN   NNPS     35
10     1     1     9        of        of    ADP     IN     43
# ... with 65,748 more rows
```

A phantom token "ROOT" is included at the start of each sentence (it always has `tid` equal to 0) if the option `include_root` is set to TRUE (it is FALSE by default). This is useful so that joins from the dependency table, which contains references to the sentence root, into the token table have no missing values.

The field upos contains the universal part of speech code, a language-agnostic classification, for the token. It could be argued that in order to maintain database normalization one should simply look up the universal part of speech code by finding the language code in the document table and joining a table mapping the Penn Treebank codes to the universal codes. This has not been done for several reasons. First, universal parts of speech are very useful for exploratory data analysis as they contain tags much more familiar to non-specialists such as "NOUN" (noun) and "CONJ" (conjunction). Asking users to apply a three table join just to access them seems overly cumbersome. Secondly, it is possible for users to use other parsers or annotation engines. These may not include granular part of speech codes and it would be difficult to figure out how to represent these if there were not a dedicated universal part of speech field.

### Dependencies

Dependencies give the grammatical relationship between pairs of tokens within a sentence (Green et al., 2011; Rafferty and Manning, 2008). As they are at the level of token pairs, they must be represented as a new table. All included fields are described in Table 4. Only one dependency should exist for any pair of tokens; the document id, sentence id, and source and target token ids together serve as a composite key. As dependencies exist only within a sentence, the sentence id does not need to be defined separately for the source and target. Dependencies take significantly longer to calculate than the lemmatization and part of speech tagging tasks.

| | get_dependency() |
|---|---|
| id | integer. Id of the source document. |
| sid | integer. Sentence id of the source token. |
| tid | integer. Id of the source token. |
| sid_target | integer. Sentence id of the target token. |
| tid_target | integer. Id of the target token. |
| relation | character. Language-agnostic universal dependency type. |
| relation_full | character. Language specific universal dependency type. |
| word | character. The source word in the raw text. |
| lemma | character. Lemmatized form of the source word. |
| word_target | character. The target word in the raw text. |
| lemma_target | character. Lemmatized form of the target word. |

**Table 4:** Schema for the dependency table. The final four variables are only provided when the option `get_token` is set to `TRUE`. The first five fields together create a composite key for the table.

The `get_dependency` function has an option (set to `FALSE` by default) to auto join the dependency to the target and source words and lemmas from the token table. This is a common task and involves non-trivial calls to the `left_join` function making it worthwhile to include as an option. For example, the following code replicates the behavior of `get_dependency` when set to return words and lemmas:

```
dep <- get_dependency(obama) %>%
  left_join(select(get_token(obama, include_root = TRUE),
                   id, sid, tid, word, lemma),
            by = c("id", "sid", "tid")) %>%
  left_join(select(get_token(obama, include_root = TRUE),
                   id, sid, tid_target = tid,
                   word_target = word, lemma_target = lemma),
            by = c("id", "sid", "tid_target"))
```

The output, equivalently using a call to `get_dependency`, is given by:

```
> get_dependency(obama, get_token = TRUE)
# A tibble: 62,781 x 10
      id   sid   tid tid_target relation relation_full    word    lemma
   <int> <int> <int>      <int>    <chr>         <chr>   <chr>    <chr>
1      1     1     2          1 compound          <NA> Speaker  speaker
2      1     1     0          2     ROOT          <NA>    ROOT     ROOT
3      1     1     2          3    punct          <NA> Speaker  speaker
4      1     1     6          4 compound          <NA> President president
5      1     1     6          5 compound          <NA> President president
6      1     1     2          6    appos          <NA> Speaker  speaker
7      1     1     6          7    punct          <NA> President president
8      1     1     6          8    appos          <NA> President president
9      1     1     8          9     prep          <NA> Members  members
10     1     1     9         10     pobj          <NA>      of       of
   word_target lemma_target
         <chr>        <chr>
1       Madam        madam
2     Speaker      speaker
3           ,            ,
4         Mr.          mr.
5        Vice         vice
6   President    president
7           ,            ,
8     Members      members
9          of           of
10   Congress     congress
# ... with 62,771 more rows
```

The word "ROOT" shows up in the first row, which would have been `NA` had sentence roots not been explicitly included in the token table.

| | get_entity() |
|---|---|
| id | integer. Id of the source document. |
| sid | integer. Sentence id of the entity mention. |
| tid | integer. Token id at the start of the entity mention. |
| tid_end | integer. Token id at the end of the entity mention. |
| entity_type | character. Type of entity. |
| entity | character. Raw words of the named entity in the text. |
| entity_normalized | character. Normalized version of the entity. |

**Table 5:** Schema for the entity table. The first three fields serve as a composite key.

Our parser produces universal dependencies (De Marneffe et al., 2014), which have a language-agnostic set of relationship types with language-specific subsets pertaining to specific grammatical relationships with a particular language. For the same reasons that both the part of speech codes and universal part of speech codes are included, each of these relationship types have been added to the dependency table.

### Named entities

*Named entity recognition* is the task of finding entities that can be defined by proper names, categorizing them, and standardizing their formats (Finkel et al., 2005). The XML output of the Stanford CoreNLP pipeline places named entity information directly into their version of the token table. Doing this repeats information over every token in an entity and gives no canonical way of extracting the entirety of a single entity mention. We instead have a separate entity table, as is demanded by the normalized database structure, and record each entity mention in its own row. The full set of fields are given in Table 5, with the combination of document id, sentence id, and token id serving as a composite key.

An example of the named entity table is given by:

```
> get_entity(obama)
# A tibble: 3,035 x 6
      id   sid   tid tid_end entity_type              entity
   <int> <int> <int>   <int>        <chr>               <chr>
1      1     1     1       2       PERSON       Madam Speaker
2      1     1     8      10          ORG Members of Congress
3      1     1    12      14          ORG      the First Lady
4      1     1    16      18          GPE   the United States
5      1     1    30      30         TIME             tonight
6      1     1    43      44        EVENT            Chamber,
7      1     2     6       6         NORP           Americans
8      1     4    24      25         DATE           every day
9      1     8    23      23         TIME             tonight
10     1     8    27      27         NORP            American
# ... with 3,025 more rows
```

The categories available in the field `entity_type` are dependent on the specific back end used. When using the coreNLP back end, the entities 'MONEY', 'ORDINAL' 'PERCENT', 'DATE' and 'TIME' also have a normalized form. Entities for the spaCy backend offer more granular distinctions, with a full list contained in the help page for the function `get_entity`. As with the coreference table, a complete representation of the entity is given as a character string due to the difficulty in reconstructing this after the fact from the token table, so the character string has been included as an explicit field.

### Coreference

Coreferences link sets of tokens that refer to the same underlying person, object, or idea (Recasens et al., 2013; Lee et al., 2013, 2011; Raghunathan et al., 2010). One common example is the linking of a noun in one sentence to a pronoun in the next sentence. The coreference table describes these relationships but is not strictly a table of coreferences. Instead, each row represents a single mention of an expression and gives a reference id indicating all of the other mentions that it also coreferences. Table 6 gives the entire schema of the coreference table. The document, reference, and mention ids serve as a composite key for the table. Links back into the token table for the start, end and head of the mention are given as well; these are pushed to the right of the table as they should be considered foreign keys within this table.

| | get_coreference() |
|---|---|
| id | integer. Id of the source document. |
| rid | integer. Relation ID. |
| mid | integer. Mention ID; unique to each coreference within a document. |
| mention | character. The mention as raw words from the text. |
| mention_type | character. One of "LIST", "NOMINAL", "PRONOMINAL", or "PROPER". |
| number | character. One of "PLURAL", "SINGULAR", or "UNKNOWN". |
| gender | character. One of "FEMALE", "MALE", "NEUTRAL", or "UNKNOWN". |
| animacy | character. One of "ANIMATE", "INANIMATE", or "UNKNOWN". |
| sid | integer. Sentence id of the coreference. |
| tid | integer. Token id at the start of the coreference. |
| tid_end | integer. Token id at the start of the coreference. |
| tid_head | integer. Token id of the head of the coreference. |

**Table 6:** Schema for the coreference table. Each row is best thought of as a coreference mention, rather than the coreference itself.

An example helps to explain exactly what the coreference table represents:

```
> get_coreference(obama)
# A tibble: 6,982 x 12
      id   rid   mid                         mention mention_type   number  gender
   <int> <int> <int>                           <chr>        <chr>    <chr>   <chr>
1      1  2049     7               the United States       PROPER SINGULAR NEUTRAL
2      1  2049    77 the United States of America       PROPER SINGULAR NEUTRAL
3      1  2049   102                         America       PROPER SINGULAR NEUTRAL
4      1  2049   315                         America       PROPER SINGULAR NEUTRAL
5      1  2049   742                      America 's       PROPER SINGULAR NEUTRAL
6      1  2049   782                         America       PROPER SINGULAR NEUTRAL
7      1  2049   939                         America       PROPER SINGULAR NEUTRAL
8      1  2049   991                         America       PROPER SINGULAR NEUTRAL
9      1  2049  1003                         America       PROPER SINGULAR NEUTRAL
10     1  2049  1045                         America       PROPER SINGULAR NEUTRAL
     animacy   sid   tid tid_end tid_head
       <chr> <dbl> <int>   <int>    <int>
1  INANIMATE     1    16      18       18
2  INANIMATE     8    41      45       43
3  INANIMATE    12     6       6        6
4  INANIMATE    40    12      12       12
5  INANIMATE   103     8       9        8
6  INANIMATE   109     8       8        8
7  INANIMATE   132     5       5        5
8  INANIMATE   138    27      27       27
9  INANIMATE   140    41      41       41
10 INANIMATE   147     4       4        4
# ... with 6,972 more rows
```

Here, these are all mentions of the same underlying entity: The United States of America. There is a special relationship between the reference id `rid` and the mention id `mid`. The coreference annotator selects a specific mention for each reference that gets treated as the canonical mention for the entire class. The mention id for this mention becomes the reference id for the class. This relationship provides a way of identifying the canonical mention within a reference class and a way of treating the coreference table as pairs of mentions rather than individual mentions joined by a given key.

The text of the mention itself is included within the table. This was done because as the mention may span several tokens it would otherwise be very difficult to extract this information from the token table. It is also possible, though not supported in the current CoreNLP pipeline, that a mention could consist of a set of non-contiguous tokens, making this field impossible to otherwise reconstruct.

| | get_sentence() |
|---|---|
| id | integer. Id of the source document. |
| sid | integer. Sentence id. |
| sentiment | integer. Predicted sentiment; 0 (very negative) to 4 (very positive). |

**Table 7:** Schema for the setence table. The document and sentence ids serve as a composite key.

### Sentence level annotations

The sentiment tagger provided by the CoreNLP pipeline predicts whether a sentence is very negative (0), negative (1), neutral (2), positive (3), or very positive (4) (Socher et al., 2013). There is no native sentiment model currently supported by spaCy. The sentiment output is placed in a separate table because it returns information exclusively at the sentence level, unlike any of the other parsers. The schema, described in Table 7, has the document and sentence ids serving as composite keys, with the only other field being an integer sentiment code. An example of the output can be seen in:

```
> get_sentence(obama)
# A tibble: 2,988 x 3
      id   sid sentiment
   <int> <dbl>     <int>
1      1     1         1
2      1     2         3
3      1     3         1
4      1     4         1
5      1     5         2
6      1     6         1
7      1     7         3
8      1     8         1
9      1     9         1
10     1    10         1
# ... with 2,978 more rows
```

The underlying sentiment model is a neural network. While at the moment few annotators exist at the sentence level, there is currently active research in modeling features that would eventually fit well into this table such as indicators of mood (Gaikwad and Joshi, 2016), levels of sarcasm (Schifanella et al., 2016) or a characterization of the sentence's "style" (Kabbara and Cheung, 2016).

### Word vectors

Our final table in the data model stores the relatively new concept of a word vector. Also known as word embeddings, these vectors are deterministic maps from the set of all available words into a high-dimensional, real valued vector space. Words with similar meanings or themes will tend to be clustered together in this high-dimensional space. For example, we would expect apple and pear to be very close to one another, with vegetables such as carrots, broccoli, and asparagus only slightly farther away. The embeddings can often be used as input features when building models on top of textual data. For a more detailed description of these embeddings, see the papers on either of the most well-known examples: GloVe (Pennington et al., 2014) and word2vec (Mikolov et al., 2013). Only the spaCy back end to **cleanNLP** currently supports word vectors; these are turned off by default because they take a significantly large amount of space to store. The embedding model uses the fasttext embeddings (Bojanowski et al., 2016), a modification of the GloVe embeddings, which map words into a 300-dimensional space. To compute the embeddings, set the vector_flag parameter of init_spaCy to TRUE prior to running the annotation.

Word vectors are stored in a separate table from the tokens table out of convenience rather than as a necessity of preserving the data model's normalized schema. Due to its size and the fact that the individual components of the word embedding have no intrinsic meaning, this table is stored as a matrix. We can see that there is exactly one row in the word embeddings for every non-ROOT token in the token table (note that the word embeddings for the obama dataset are not included with the package as they are too large to be uploaded to CRAN).

```
> dim(get_token(obama))
[1] 62781     8
> dim(get_vector(obama))
[1] 62781   303
```

The first three columns hold the keys `id`, `sid`, and `tid`, respectively. If no embedding is computed, the function `get_vector` returns an empty matrix.

## Using cleanNLP to study State of the Union addresses

The President of the United States is constitutionally obligated to provide a report known as the *State of the Union*. The report summarizes the current challenges facing the country and the president's upcoming legislative agenda. While historically the State of the Union was often a written document, in recent decades it has always taken the form of an oral address to a joint session of the United States Congress. In this final section the utility of the package is illustrated by showing how it can be used to study a corpus consisting of every such address made by a United States president through 2016 (Peters, 2016). It highlights some of the major benefits of the tidy data model as it applies to the study of textual data, though by no means attempts to give an exhaustive coverage of all the available tables and approaches. The examples make heavy use of the table verbs provided by **dplyr**, the piping notation of **magrittr** and **ggplot2** graphics. These are used because they best illustrate the advantages of the tidy data model that has been built in **cleanNLP** for representing corpus annotations. Relevant functions are prepended with `cleanNLP::` in the following analysis in order to be clear which functions are supplied by the **cleanNLP** package.

### Loading and parsing the data

The full text of all the State of the Union addresses through 2016 are available in the R package sotu (Arnold, 2017), available on CRAN. The package also contains meta-data concerning each speech that we will add to the document table while annotating the corpus. The code to run this annotation is given by:

```
> library(sotu)
> library(cleanNLP)
>
> data(sotu_text)
> data(sotu_meta)
> init_spaCy()
> sotu <- cleanNLP::run_annotators(sotu_text, as_strings = TRUE,
+                                  meta = sotu_meta)
```

The annotation object, which we will use in the example in the following analysis, is stored in the object sotu.

### Exploratory analysis

Simple summary statistics are easily computed off of the token table. To see the distribution of sentence length, the token table is grouped by the document and sentence id and the number of rows within each group are computed. The percentiles of these counts give a quick summary of the distribution.

```
> library(ggplot2)
> library(dplyr)
> cleanNLP::get_token(sotu) %>%
+   count(id, sid) %$%
+   quantile(n, seq(0,1,0.1))
  0%  10%  20%  30%  40%  50%  60%  70%  80%  90% 100%
   1   11   16   19   23   27   31   37   44   58  681
```

The median sentence has 28 tokens, whereas at least one has over 600 (this is due to a bulleted list in one of the written addresses being treated as a single sentence) To see the most frequently used nouns in the dataset, the token table is filtered on the universal part of speech field, grouped by lemma, and the number of rows in each group are once again calculated. Sorting the output and selecting the top 42 nouns, yields a high level summary of the topics of interest within this corpus.

```
> cleanNLP::get_token(sotu) %>%
+   filter(upos == "NOUN") %>%
+   count(lemma) %>%
+   top_n(n = 42, n) %>%
+   arrange(desc(n)) %>%
+   use_series(lemma)
```

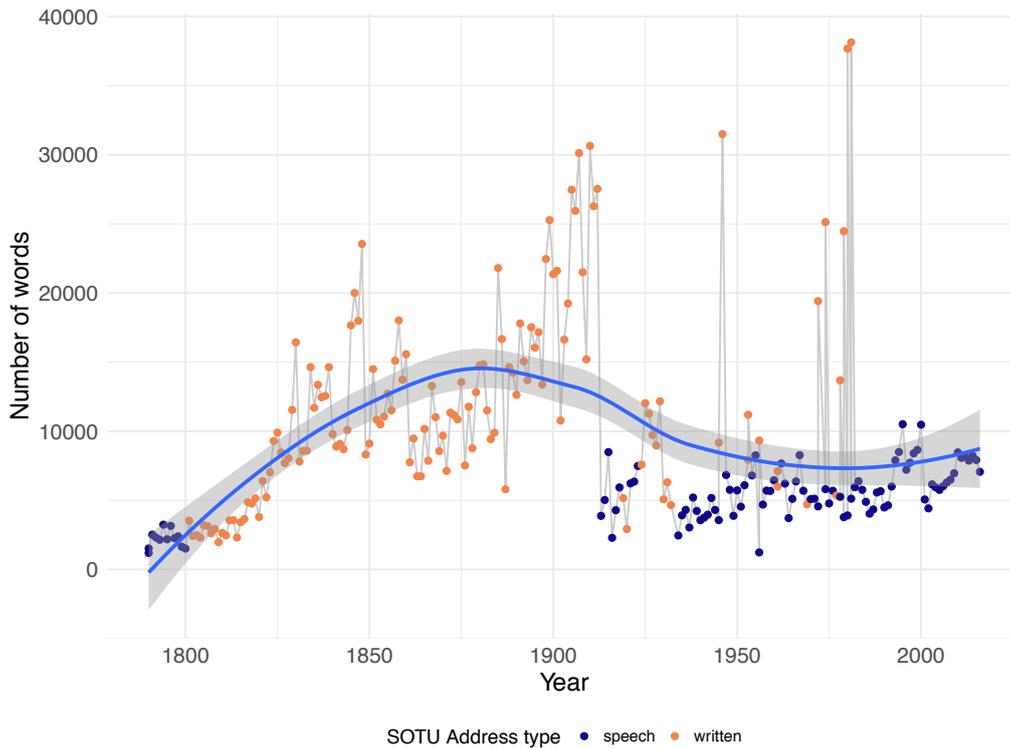

**Figure 1:** Length of each State of the Union address, in total number of tokens. Color shows whether the address was given as a speech or delivered as a written document.

```
 [1] "year"        "country"      "people"       "government"
 [5] "law"         "time"         "nation"       "who"
 [9] "power"       "interest"     "world"        "war"
[13] "citizen"     "service"      "duty"         "part"
[17] "system"      "peace"        "right"        "man"
[21] "program"     "policy"       "work"         "act"
[25] "state"       "condition"    "subject"      "legislation"
[29] "force"       "effort"       "treaty"       "purpose"
[33] "what"        "land"         "business"     "action"
[37] "measure"     "tax"          "way"          "question"
[41] "relation"    "consideration"
```

The result is generally as would be expected from a corpus of government speeches, with references to proper nouns representing various organizations within the government and non-proper nouns indicating general topics of interest such as "tax", "law", and "peace".

The length in tokens of each address is calculated similarly by grouping and summarizing at the document id level. The results can be joined with the document table to get the year of the speech and then piped in a **ggplot2** command to illustrate how the length of the State of the Union has changed over time.

```
> cleanNLP::get_token(sotu) %>%
+   count(id) %>%
+   left_join(cleanNLP::get_document(sotu)) %>%
+   ggplot(aes(year, n)) +
+     geom_line(color = grey(0.8)) +
+     geom_point(aes(color = sotu_type)) +
+     geom_smooth()
```

Here, color is used to represent whether the address was given as an oral address or a written document. The output in Figure 1 shows that their are certainly time trends to the address length, with the form of the address (written versus spoken) also having a large effect on document length.

Finding the most used entities from the entity table over the time period of the corpus yields an alternative way to see the underlying topics. A slightly modified version of the code snippet used to find the top nouns in the dataset can be used to find the top entities. The `get_token` function is replaced by `get_entity` and the table is filtered on `entity_type` rather than the universal part of speech code.

```
> cleanNLP::get_entity(sotu) %>%
+   filter(entity_type == "GPE") %>%
+   count(entity) %>%
+   top_n(n = 26, n) %>%
+   arrange(desc(n)) %>%
+   use_series(entity)
 [1] "the United States"       "America"
 [3] "States"                  "Mexico"
 [5] "Great Britain"           "Spain"
 [7] "Washington"              "China"
 [9] "Executive"               "France"
[11] "Cuba"                    "Japan"
[13] "Texas"                   "Russia"
[15] "The United States"       "Germany"
[17] "United States"           "California"
[19] "Nicaragua"               "the Soviet Union"
[21] "Mississippi"             "Iraq"
[23] "Alaska"                  "U.S."
[25] "Philippines"             "Panama"
[27] "the District of Columbia"
```

The ability to redo analyses from a slightly different perspective is a direct consequence of the tidy data model supplied by **cleanNLP**. The top locations include some obvious and some less obvious instances. Those sovereign nations included such as Great Britain, Mexico, Germany, and Japan seem as expected given either the United State's close ties or periods of war with them. The top states include the most populous regions (New York, California, and Texas) but also smaller states (Kansas, Oregon, Mississippi), the latter being more surprising.

One of the most straightforward way of extracting a high-level summary of the content of a speech is to extract all direct object dependencies where the target noun is not a very common word. In order to do this for a particular speech, the dependency table is joined to the document table, a particular document is selected, and relationships of type "dobj" (direct object) are filtered out. The result is then joined to the data set `word_frequency`, which is included with **cleanNLP**, and pairs with a target occurring less than 0.5% of the time are selected to give the final result. Here is an example of this using the first address made by George W. Bush in 2001:

```
> cleanNLP::get_dependency(sotu, get_token = TRUE) %>%
+   left_join(get_document(sotu)) %>%
+   filter(year == 2001, relation == "dobj") %>%
+   select(id = id, start = word, word = lemma_target) %>%
+   left_join(word_frequency) %>%
+   filter(frequency < 0.001) %>%
+   select(id, start, word) %$%
+   sprintf("%s => %s", start, word)
Joining, by = "id"
Joining, by = "word"
 [1] "take => oath"              "using => statistic"
 [3] "increasing => layoff"      "protects => trillion"
 [5] "makes => welcoming"        "accelerating => cleanup"
 [7] "fight => homelessness"     "helping => neighbor"
 [9] "allowing => taxpayer"      "provide => mentor"
[11] "fight => illiteracy"       "promotes => compassion"
[13] "asked => ashcroft"         "end => profiling"
[15] "pay => trillion"           "throw => dart"
[17] "restores => fairness"      "promoting => internationalism"
[19] "makes => downpayment"      "discard => relic"
[21] "confronting => shortage"   "directed => cheney"
[23] "sound => footing"          "divided => conscience"
[25] "done => servant"
```

Most of these phrases correspond with the "compassionate conservatism" that George W. Bush ran under in the preceding 2000 election. Applying the same analysis to the 2002 State of the Union, which came under the shadow of the September 11th terrorist attacks, shows a drastic shift in focus.

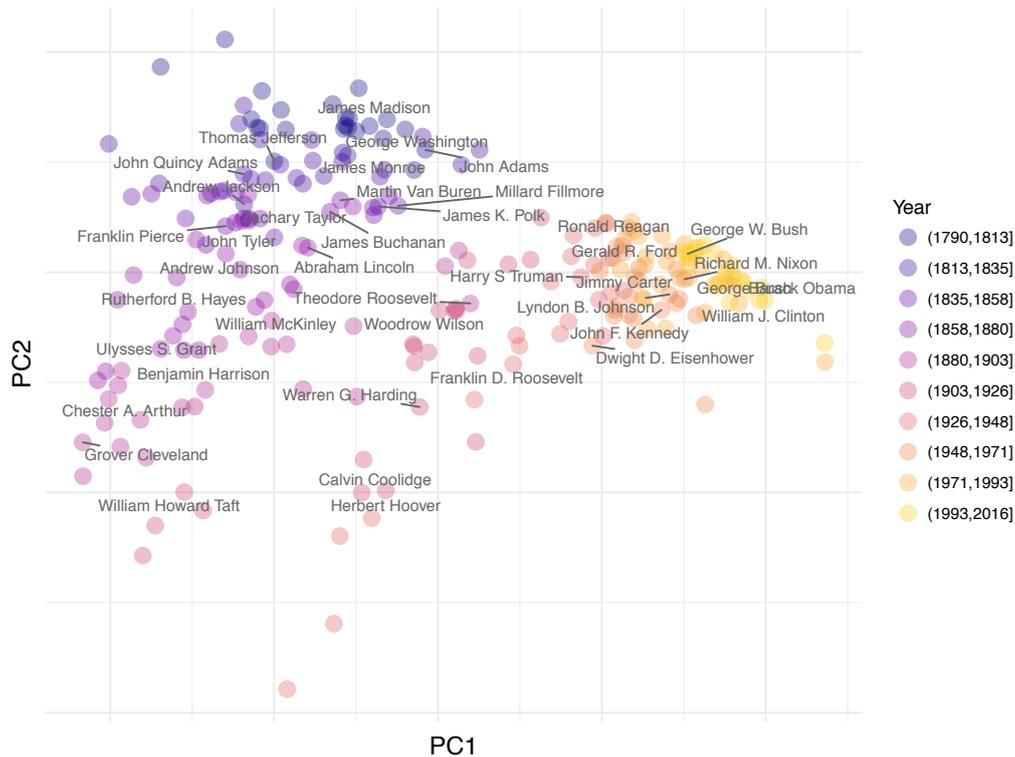

**Figure 2:** State of the Union Speeches, highlighting each President's first address, plotted using the first two principal components of the term frequency matrix of non-proper nouns.

```
> cleanNLP::get_dependency(sotu, get_token = TRUE) %>%
+    left_join(get_document(sotu)) %>%
+    filter(year == 2002, relation == "dobj") %>%
+    select(id = id, start = word, word = lemma_target) %>%
+    left_join(word_frequency) %>%
+    filter(frequency < 0.0005) %>%
+    select(id, start, word) %$%
+    sprintf("%s => %s", start, word)
Joining, by = "id"
Joining, by = "word"
 [1] "urged => follower"       "called => troop"
 [3] "brought => sorrow"       "owe => micheal"
 [5] "ticking => timebomb"     "have => troop"
 [7] "hold => hostage"         "eliminate => parasite"
 [9] "flaunt => hostility"     "develop => anthrax"
[11] "put => troop"            "increased => vigilance"
[13] "fight => anthrax"        "thank => attendant"
[15] "defeat => recession"     "want => paycheck"
[17] "set => posturing"        "enact => safeguard"
[19] "embracing => ethic"      "owns => aspiration"
[21] "containing => resentment" "erasing => rivalry"
[23] "embrace => tyranny"
```

Here the topics have almost entirely shifted to counter-terrorism and national security efforts.

## Models

The get_tfidf function provided by **cleanNLP** converts a token table into a sparse matrix representing the term-frequency inverse document frequency matrix (or any intermediate part of that calculation). This is particularly useful when building models from a textual corpus. The tidy_pca, also included with the package, takes a matrix and returns a data frame containing the desired number of principal components. Dimension reduction involves piping the token table for a corpus into the get_tfidf

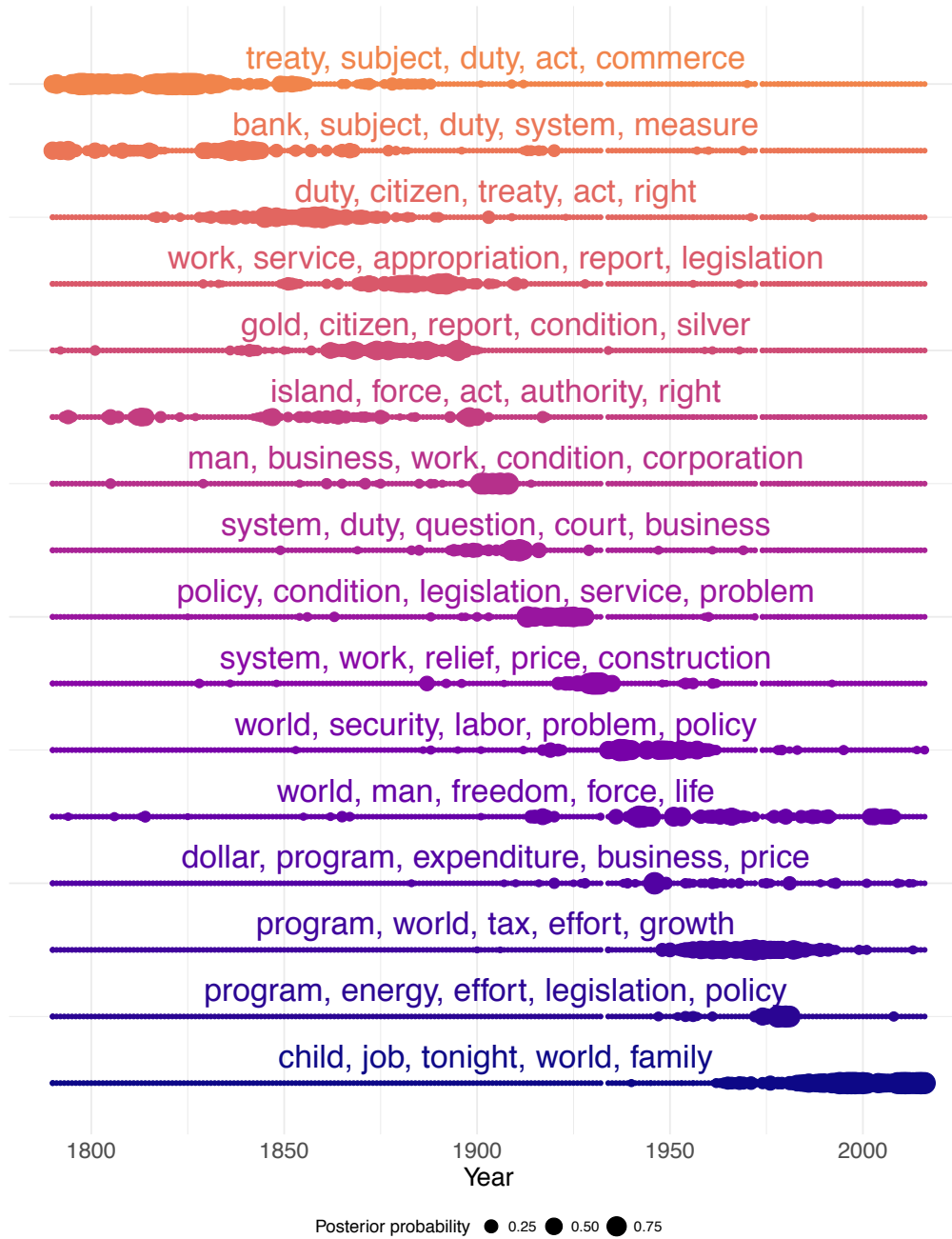

**Figure 3:** Distribution of topic model posterior probabilities over time on the State of the Union corpus. The top five words associated with each topic are displayed, with topics sorted by the median year of all documents placed into the respective topic using the maximum posterior probabilities.

function and passing the results to `tidy_pca`.

```
> pca <- cleanNLP::get_token(sotu) %>%
+   filter(pos %in% c("NN", "NNS")) %>%
+   cleanNLP::get_tfidf(min_df = 0.05, max_df = 0.95,
+                       type = "tfidf", tf_weight = "dnorm") %$%
+   cleanNLP::tidy_pca(tfidf, get_document(sotu))
```

In this example only non-proper nouns have been included in order to minimize the stylistic attributes of the speeches in order to focus more on their content. A scatter plot of the speeches using these components is shown in Figure 2. There is a definitive temporal pattern to the documents, with the 20th century addresses forming a distinct cluster on the right side of the plot.

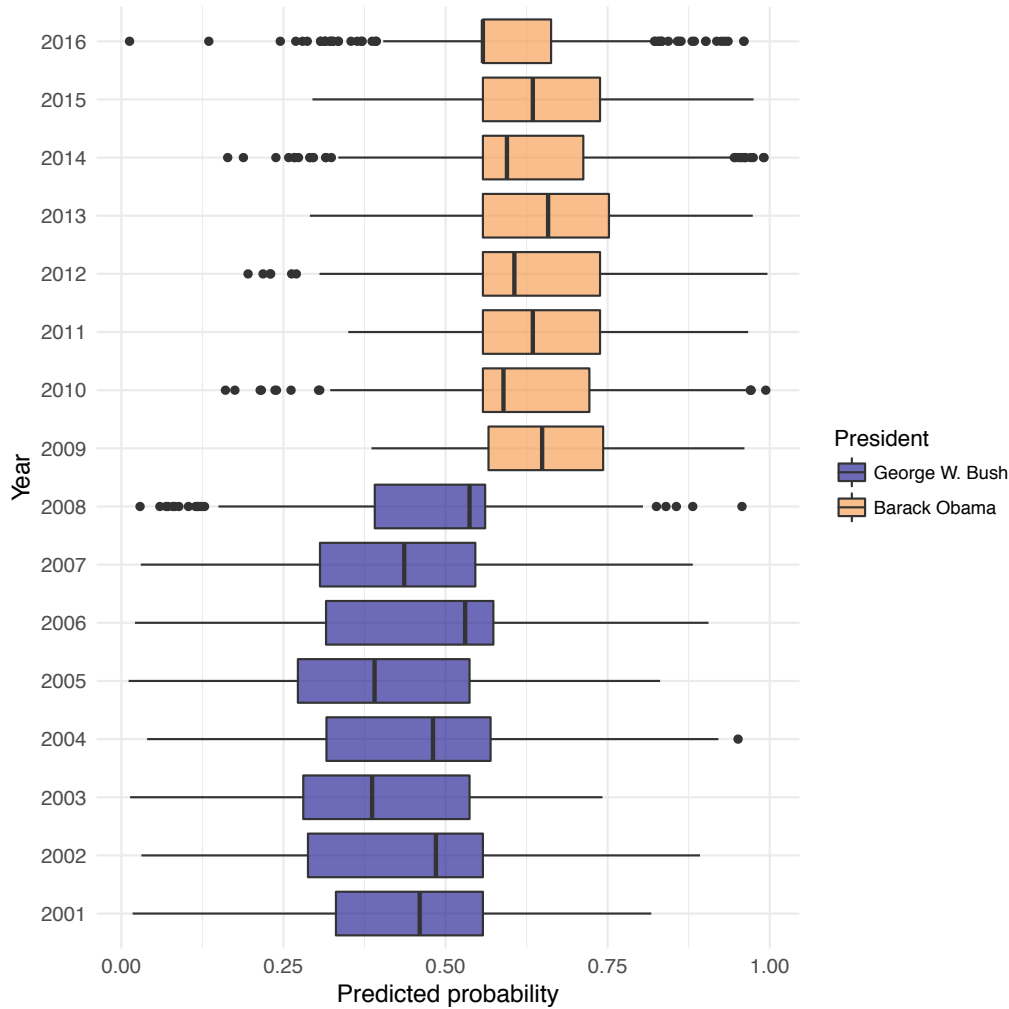

**Figure 4:** Boxplot of predicted probabilities, at the sentence level, for all 16 State of the Union addresses by Presidents George W. Bush and Barack Obama. The probability represents the extent to which the model believe the sentence was spoken by President Obama. Odd years were used for training and even years for testing. Cross-validation on the training set was used, with the one standard error rule, to set the lambda tuning parameter.

Topic models are a collection of statistical models for describing abstract themes within a textual corpus. Each theme is characterized by a collection of words that commonly co-occur; for example, the words 'crop', 'dairy', 'tractor', and 'hectare', might define a *farming* theme. One of the most popular topic models is latent Dirichlet allocation (LDA), a Bayesian model where each topic is described by a probability distribution over a vocabulary of words. Each document is then characterized by a probability distribution over the available topics. For a formal description, see Blei et al. (2003) and Pritchard et al. (2000), the original papers outlining LDA. To fit LDA on a corpus of text parsed by the **cleanNLP** package, the output of get_tfidf can be piped directly to the LDA function in the package **topicmodels**. The topic model function requires raw counts, so the type variable in get_tfidf is set to "tf".

```
> library(topicmodels)
> tm <- cleanNLP::get_token(sotu) %>%
+   filter(pos %in% c("NN", "NNS")) %>%
+   cleanNLP::get_tfidf(min_df = 0.05, max_df = 0.95,
+                       type = "tf", tf_weight = "raw") %
+   LDA(tf, k = 16, control = list(verbose = 1))
```

The topics, ordered by approximate time period, are visualized in Figure 3. We describe each topic by giving the five most important words Most topics exist for a few decades and then largely disappear, though some persist over non-contiguous periods of the presidency. The "program, energy, effort, legislation, policy" topic, for example, appears during the 1950s and crops up again during the energy crisis of the 1970s. The "world, man, freedom, force, life" topic peaks during both World Wars, but is absent during the 1920s and early 1930s.

Finally, the **cleanNLP** data model is also convenient for building predictive models. The State of the Union corpus does not lend itself to an obviously applicable prediction problem. A classifier that distinguishes speeches made by George W. Bush and Barrack Obama will be constructed here for the purpose of illustration. As a first step, a term-frequency matrix is extracted using the same technique as was used with the topic modeling function. However, here the frequency is computed for each sentence in the corpus rather than the document as a whole. The ability to do this seamlessly with a single additional `mutate` function defining a new id illustrates the flexibility of the `get_tfidf` function.

```
> df <- get_token(sotu) %>%
+    left_join(get_document(sotu)) %>%
+    filter(year > 2000) %>%
+    mutate(new_id = paste(id, sid, sep = "-")) %>%
+    filter(pos %in% c("NN", "NNS"))
Joining, by = "id"
> mat <- get_tfidf(df, min_df = 0, max_df = 1, type = "tf",
+                  tf_weight = "raw", doc_var = "new_id")
```

It will be nessisary to define a response variable y indicating whether this is a speech made by President Obama as well as a training flag indicating which speeches were made in odd numbered years. This is done via a separate table join and a pair of mutations.

```
> meta <- data_frame(new_id = mat$id) %>%
+    left_join(df[!duplicated(df$new_id),]) %>%
+    mutate(y = as.numeric(president == "Barack Obama")) %>%
+    mutate(train = year %in% seq(2001,2016, by = 2))
Joining, by = "new_id"
```

The output may now be used as input to the elastic net function provided by the **glmnet** package. The response is set to the binomial family given the binary nature of the response and training is done on only those speeches occurring in odd-numbered years. Cross-validation is used in order to select the best value of the model's tuning parameter.

```
> library(glmnet)
> model <- cv.glmnet(mat$tf[meta$train,], meta$y[meta$train], family = "binomial")
```

A boxplot of the predicted classes for each address is given in Figure 4. The algorithm does a very good job of separating the speeches. Looking at the odd years versus even years (the training and testing sets, respectively) indicates that the model has not been over-fit.

One benefit of the penalized linear regression model is that it is possible to interpret the coefficients in a meaningful way. Here are the non-zero elements of the regression vector, coded as whether the have a positive (more Obama) or negative (more Bush) sign:

```
> beta <- coef(model, s = model[["lambda"]][11])[-1]
> sprintf("%s (%d)", mat$vocab, sign(beta))[beta != 0]
 [1] "job (1)"         "business (1)"     "citizen (-1)"
 [4] "terrorist (-1)"  "government (-1)"  "freedom (-1)"
 [7] "home (1)"        "college (1)"      "weapon (-1)"
[10] "deficit (1)"     "company (1)"      "peace (-1)"
[13] "enemy (-1)"      "terror (-1)"      "income (-1)"
[16] "drug (-1)"       "kid (1)"          "regime (-1)"
[19] "class (1)"       "crisis (1)"       "industry (1)"
[22] "need (-1)"       "fact (1)"         "relief (-1)"
[25] "bank (1)"        "liberty (-1)"     "society (-1)"
[28] "duty (-1)"       "folk (1)"         "account (-1)"
[31] "compassion (-1)" "environment (-1)" "inspector (-1)"
```

These generally seem as expected given the main policy topics of focus under each administration. During most of the Bush presidency, as mentioned before, the focus was on national security and foreign policy. Obama, on the other hand, inherited the recession of 2008 and was more focused on the overall economic policy.

## Conclusions

In this paper a normalized data model for representing text annotations has been presented and rationalized. We have also demonstrated how the R package **cleanNLP** implements this data model using various, configurable back ends. Our focus has been to illustrate why this general approach and specific implementation is both powerful and easy to integrate into existing data pipelines. It is expected that some users will utilize the entirety of the underlying annotation pipelines, internal R structures, and helper functions. Others may use the package as a convenient wrapper around either the CoreNLP or spaCy libraries. In either extreme, or anywhere in between, our approach provides powerful tools for applying exploratory, graphical, and model-based techniques to textual data sources.

The **cleanNLP** package continues to be actively developed. In particular, we hope to include new sentence-level annotations as they are integrated into the spaCy and CoreNLP libraries. While major releases are available on CRAN, new features are added periodically on the development branch located at: https://github.com/statsmaths/cleanNLP. Bug reports, feature and collaboration requests can all be made using the GitHub issues page.

## Bibliography

J. Allaire, K. Ushey, Y. Tang, and D. Eddelbuettel. *reticulate: R Interface to Python*, 2017. URL https://github.com/rstudio/reticulate. [p249]

T. Arnold and L. Tilton. *coreNLP: Wrappers Around Stanford CoreNLP Tools*, 2016. R package version 0.4-2. [p249]

T. B. Arnold. *sotu: United States Presidential State of the Union Addresses*, 2017. URL https://CRAN.R-project.org/package=sotu. R package version 1.0.2. [p258]

S. M. Bache and H. Wickham. *magrittr: A Forward-Pipe Operator for R*, 2014. URL https://CRAN.R-project.org/package=magrittr. R package version 1.5. [p248]

J. Baldridge. The openNLP project. *URL http://opennlp.apache.org/index.html,(accessed 2 February 2012)*, 2005. [p248]

K. Benoit and A. Matsuo. *spacyr: R Wrapper to the spaCy NLP Library*, 2017. URL http://github.com/kbenoit/spacyr. R package version 0.9.0. [p249]

D. M. Blei, A. Y. Ng, and M. I. Jordan. Latent dirichlet allocation. *Journal of machine Learning research*, 3 (Jan):993–1022, 2003. [p263]

P. Bojanowski, E. Grave, A. Joulin, and T. Mikolov. Enriching word vectors with subword information. *arXiv preprint arXiv:1607.04606*, 2016. [p257]

S. Buchholz and E. Marsi. Conll-x shared task on multilingual dependency parsing. In *Proceedings of the Tenth Conference on Computational Natural Language Learning*, pages 149–164. Association for Computational Linguistics, 2006. [p252]

J. Chang. *lda: Collapsed Gibbs Sampling Methods for Topic Models*, 2015. URL https://CRAN.R-project.org/package=lda. R package version 1.4.2. [p249]

E. F. Codd. *The relational model for database management: version 2*. Addison-Wesley Longman Publishing Co., Inc., 1990. [p251]

M.-C. De Marneffe, T. Dozat, N. Silveira, K. Haverinen, F. Ginter, J. Nivre, and C. D. Manning. Universal Stanford dependencies: A cross-linguistic typology. In *LREC*, volume 14, pages 4585–92, 2014. [p255]

I. Feinerer, K. Hornik, and D. Meyer. Text mining infrastructure in R. *Journal of statistical software*, 25(5): 1–54, 2008. URL https://doi.org/10.18637/jss.v025.i05. [p249]

J. R. Finkel, T. Grenager, and C. Manning. Incorporating non-local information into information extraction systems by Gibbs sampling. In *Proceedings of the 43rd Annual Meeting on Association for Computational Linguistics*, pages 363–370. Association for Computational Linguistics, 2005. [p255]

S. Firke. *janitor: Simple Tools for Examining and Cleaning Dirty Data*, 2016. URL https://CRAN.R-project.org/package=janitor. R package version 0.2.1. [p248]

W. Freitas. *sqliter: Connection wrapper to SQLite databases*, 2014. URL https://CRAN.R-project.org/package=sqliter. R package version 0.1.0. [p249]

G. Gaikwad and D. J. Joshi. Multiclass mood classification on twitter using lexicon dictionary and machine learning algorithms. In *Inventive Computation Technologies (ICICT), International Conference on*, volume 1, pages 1–6. IEEE, 2016. [p257]

S. Green, M.-C. De Marneffe, J. Bauer, and C. D. Manning. Multiword expression identification with tree substitution grammars: A parsing tour de force with french. In *Proceedings of the Conference on Empirical Methods in Natural Language Processing*, pages 725–735. Association for Computational Linguistics, 2011. [p253]

B. Grün and K. Hornik. topicmodels: An R package for fitting topic models. *Journal of Statistical Software*, 40(13):1–30, 2011. URL https://doi.org/10.18637/jss.v040.i13. [p249]

S. Hellmann, J. Lehmann, and S. Auer. Nif: An ontology-based and linked-data-aware NLP interchange format. *Working Draft*, 2012. [p252]

M. Honnibal and M. Johnson. An improved non-monotonic transition system for dependency parsing. In *Proceedings of the 2015 Conference on Empirical Methods in Natural Language Processing*, pages 1373–1378, Lisbon, Portugal, September 2015. Association for Computational Linguistics. URL https://aclweb.org/anthology/D/D15/D15-1162. [p248]

K. Hornik. *NLP: Natural Language Processing Infrastructure*, 2016a. URL https://CRAN.R-project.org/package=NLP. R package version 0.1-9. [p249]

K. Hornik. *openNLP: Apache OpenNLP Tools Interface*, 2016b. URL https://CRAN.R-project.org/package=openNLP. R package version 0.2-6. [p249]

K. Hornik. *StanfordCoreNLP*, 2016c. URL http://datacube.wu.ac.at/src/contrib/. R package version 0.1-1. [p249]

N. Ide and L. Romary. A common framework for syntactic annotation. In *Proceedings of the 39th Annual Meeting on Association for Computational Linguistics*, pages 306–313. Association for Computational Linguistics, 2001. [p252]

J. Kabbara and J. C. K. Cheung. Stylistic transfer in natural language generation systems using recurrent neural networks. *EMNLP 2016*, page 43, 2016. [p257]

H. Lee, Y. Peirsman, A. Chang, N. Chambers, M. Surdeanu, and D. Jurafsky. Stanford's multi-pass sieve coreference resolution system at the CoNLL-2011 shared task. In *Proceedings of the Fifteenth Conference on Computational Natural Language Learning: Shared Task*, pages 28–34. Association for Computational Linguistics, 2011. [p255]

H. Lee, A. Chang, Y. Peirsman, N. Chambers, M. Surdeanu, and D. Jurafsky. Deterministic coreference resolution based on entity-centric, precision-ranked rules. *Computational Linguistics*, 39(4):885–916, 2013. URL https://doi.org/10.1162/coli_a_00152. [p255]

C. D. Manning, M. Surdeanu, J. Bauer, J. R. Finkel, S. Bethard, and D. McClosky. The Stanford CoreNLP natural language processing toolkit. In *ACL (System Demonstrations)*, pages 55–60, 2014. [p248]

W. McKinney et al. Data structures for statistical computing in Python. In *Proceedings of the 9th Python in Science Conference*, volume 445, pages 51–56. van der Voort S, Millman J, 2010. [p248]

T. Mikolov, I. Sutskever, K. Chen, G. S. Corrado, and J. Dean. Distributed representations of words and phrases and their compositionality. In *Advances in neural information processing systems*, pages 3111–3119, 2013. [p257]

L. Mullen. *tokenizers: A Consistent Interface to Tokenize Natural Language Text*, 2016. URL https://CRAN.R-project.org/package=tokenizers. R package version 0.1.4. [p249]

F. Pedregosa, G. Varoquaux, A. Gramfort, V. Michel, B. Thirion, O. Grisel, M. Blondel, P. Prettenhofer, R. Weiss, V. Dubourg, et al. Scikit-learn: Machine learning in python. *Journal of Machine Learning Research*, 12(Oct):2825–2830, 2011. [p248]

J. Pennington, R. Socher, and C. D. Manning. GloVe: Global vectors for word representation. In *EMNLP*, volume 14, pages 1532–1543, 2014. URL https://doi.org/10.3115/v1/D14-1162. [p257]

G. Peters. *State of the Union Addresses and Messages*, 2016. URL http://www.presidency.ucsb.edu/sou.php. [p258]

S. Petrov. Announcing SyntaxNet: The world's most accurate parser goes open source. *Google Research Blog, May*, 12:2016, 2016. [p248]

J. K. Pritchard, M. Stephens, and P. Donnelly. Inference of population structure using multilocus genotype data. *Genetics*, 155(2):945–959, 2000. [p263]

A. N. Rafferty and C. D. Manning. Parsing three German treebanks: Lexicalized and unlexicalized baselines. In *Proceedings of the Workshop on Parsing German*, pages 40–46. Association for Computational Linguistics, 2008. [p253]

K. Raghunathan, H. Lee, S. Rangarajan, N. Chambers, M. Surdeanu, D. Jurafsky, and C. Manning. A multi-pass sieve for coreference resolution. In *Proceedings of the 2010 Conference on Empirical Methods in Natural Language Processing*, pages 492–501. Association for Computational Linguistics, 2010. [p255]

M. Recasens, M.-C. de Marneffe, and C. Potts. The life and death of discourse entities: Identifying singleton mentions. In *HLT-NAACL*, pages 627–633, 2013. [p255]

D. Robinson. *broom: Convert Statistical Analysis Objects into Tidy Data Frames*, 2017. URL https://CRAN.R-project.org/package=broom. R package version 0.4.2. [p248]

R. Schifanella, P. de Juan, J. Tetreault, and L. Cao. Detecting sarcasm in multimodal social platforms. In *Proceedings of the 2016 ACM on Multimedia Conference*, pages 1136–1145. ACM, 2016. [p257]

J. Silge and D. Robinson. tidytext: Text mining and analysis using tidy data principles in R. *The Journal of Open Source Software*, 1(3), 2016. URL https://doi.org/10.21105/joss.00037. [p249]

R. Socher, A. Perelygin, J. Y. Wu, J. Chuang, C. D. Manning, A. Y. Ng, and C. Potts. Recursive deep models for semantic compositionality over a sentiment treebank. In *Proceedings of the conference on empirical methods in natural language processing (EMNLP)*, volume 1631, page 1642. Citeseer, 2013. [p257]

K. Toutanova and C. D. Manning. Enriching the knowledge sources used in a maximum entropy part-of-speech tagger. In *Proceedings of the 2000 Joint SIGDAT conference on Empirical methods in natural language processing and very large corpora: held in conjunction with the 38th Annual Meeting of the Association for Computational Linguistics-Volume 13*, pages 63–70. Association for Computational Linguistics, 2000. [p253]

S. Urbanek. *rJava: Low-Level R to Java Interface*, 2016. URL https://CRAN.R-project.org/package=rJava. R package version 0.9-8. [p248]

H. Wickham. *ggplot2: Elegant Graphics for Data Analysis*. Springer-Verlag New York, 2009. ISBN 978-0-387-98140-6. URL https://doi.org/10.1007/978-0-387-98141-3. [p248]

H. Wickham. Tidy data. *Journal of Statistical Software*, 59(i10), 2014. URL https://doi.org/10.18637/jss.v059.i10. [p248, 251]

H. Wickham. *tidyr: Easily Tidy Data with 'spread()' and 'gather()' Functions*, 2017. URL https://CRAN.R-project.org/package=tidyr. R package version 0.6.1. [p248]

H. Wickham and R. Francois. *dplyr: A Grammar of Data Manipulation*, 2016. URL https://CRAN.R-project.org/package=dplyr. R package version 0.5.0. [p248]

F. Wild. *lsa: Latent Semantic Analysis*, 2015. URL https://CRAN.R-project.org/package=lsa. R package version 0.73.1. [p249]

*Taylor Arnold*
*Department of Mathematics and Computer Science*
*University of Richmond*
*Richmond, VA 23173 USA*
tarnold2@richmond.edu